\pdfoutput=1
\documentclass[11pt]{article}

\usepackage[final]{acl}

\usepackage{times}
\usepackage{latexsym}
\usepackage{amsmath}
\usepackage{graphicx}
\usepackage{caption}
\usepackage{tabularx}
\usepackage{multirow}
\usepackage{booktabs}
\usepackage{float}                  % 图片浮动位置
\usepackage{subfig}                 % 子图包，不要与{subfigure}混用，{subfig}较新
\usepackage{overpic}                % 叭啦叭啦，与图片排版相关
\usepackage{color}
\usepackage[normalem]{ulem}
\usepackage{array}
\usepackage{soul}
\usepackage{pifont}
\usepackage{ulem}
\usepackage{listings}
\usepackage{xcolor} % 用于定制颜色

\usepackage[T1]{fontenc}
\usepackage[utf8]{inputenc}

\usepackage{microtype}

\usepackage{inconsolata}

\usepackage{graphicx}
\usepackage{amsfonts}
\usepackage{amssymb}
\usepackage{hyperref}

\title{MARE: Multi-Aspect Rationale Extractor on Unsupervised Rationale Extraction}

\author{Han Jiang \and  Junwen Duan* \and Zhe Qu \and Jianxin Wang \\ 
Hunan Provincial Key Lab on Bioinformatics, School of Computer Science and Engineering,\\
Central South University, Changsha, Hunan, China \\
\{jh-better, jwduan, zhe\_qu\}@csu.edu.cn, jxwang@mail.csu.edu.cn
}

\begin{document}
\maketitle
\begin{abstract}
Unsupervised rationale extraction aims to extract text snippets to support model predictions without explicit rationale annotation.
Researchers have made many efforts to solve this task. 
% However, Previous works encode each aspect independently, ignoring their internal correlations.
% Meanwhile, such a uni-aspect encoding model can only explain and predict one aspect of the text at a time, which limits its downstream applications. 
Previous works often encode each aspect independently, which may limit their ability to capture meaningful internal correlations between aspects.
While there has been significant work on mitigating spurious correlations, our approach focuses on leveraging the beneficial internal correlations to improve multi-aspect rationale extraction.
In this paper, we propose a Multi-Aspect Rationale Extractor~(MARE) to explain and predict multiple aspects simultaneously.
Concretely, we propose a Multi-Aspect Multi-Head Attention~(MAMHA) mechanism based on \textit{hard deletion} to encode multiple text chunks simultaneously.
Furthermore, multiple special tokens are prepended in front of the text with each corresponding to one certain aspect.
Finally, multi-task training is deployed to reduce the training overhead.
Experimental results on two unsupervised rationale extraction benchmarks show that MARE achieves state-of-the-art performance.
Ablation studies further demonstrate the effectiveness of our method.
Our codes have been available at \url{https://github.com/CSU-NLP-Group/MARE}.
\end{abstract}

\section{Introduction}

% Deep learning methods have achieved state-of-the-art performance on various downstream tasks across numerous fields, such as computer vision and natural language processing. 
Deep learning text classification systems have achieved remarkable performance in recent years~\cite{YoonKim2014ConvolutionalNN, JacobDevlin2018BERTPO}. However, their black-box nature has been widely criticized. Finding a sufficient approach to open the black box is urgent and significant.

\begin{figure}
    \centering
    \subfloat[Typical uni-aspect encoding models]{\includegraphics[width=.45\columnwidth]{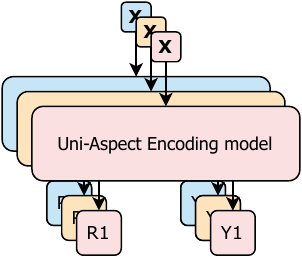}\label{fig:uni}}
    \subfloat[MARE~(\textit{ours})]{\includegraphics[width=.45\columnwidth]{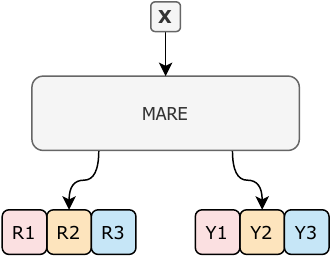}\label{fig:MARE}} 
    \caption{Comparison of our methods (MARE) with previous typical uni-aspect encoding models.% a). the basic frameworks proposed in RNP. b). The folded Rationalization model from FR shares the backbone of the generator and predictor. c). Our proposed framework, generates the rationale and prediction with a single forward pass.
     }
    % \vspace{-1em}
\end{figure}

Unsupervised rationale extraction~\cite{lei-etal-2016-rationalizing} is an explanation approach that aims to extract text snippets from input text to support model predictions without explicit rationale annotation. Previous researchers~\cite{FR, YOFO} have made many efforts to improve the rationalization performance of their models. However, as shown in Figure~\ref{fig:uni}, existing rationale extraction models are uni-aspect encoding models, which can only predict and interpret one aspect of the text at a time. In real-world scenarios, one text often contains multiple aspects of an object.
Table~\ref{table:beer_exp} shows an example from the BeerAdvocate dataset~\cite{beer}, where \textcolor{blue}{blue}, \textcolor{red}{red}, and \textcolor{cyan}{cyan} represent the aspects of \textit{Appearance}, \textit{Aroma}, and \textit{Palate}, respectively. The highlighted segments in the text are the rationales corresponding to each aspect. For instance, "\textit{pours a murky orangish-brown color with a white head .}" explains why the label for \textit{Appearance} is Positive. In this case, traditional uni-aspect rationale extraction models would require three independently trained models to predict and interpret all three aspects, which is labor-intensive and time-consuming and limits their downstream applications. Furthermore, uni-aspect models encode each aspect independently ignoring their internal correlation.

\begin{table}[]
    \centering
    \begin{tabular}{p{7cm}} 
    \toprule
    \hline
    Example\\
    \hline
    \textbf{Appearance}: \textcolor{blue}{Positive} \\
    \textbf{Aroma}: \textcolor{red}{Positive} \\
    \textbf{Palate}: \textcolor{cyan}{Positive}  \\
    \textbf{Text}: thanks to bman1113vr for sharing this bottle . \textcolor{blue}{pours a murky orangish-brown color with a white head .}  \textcolor{red}{the aroma   is tart lemons .} the flavor is tart lemons with some oak-aged character . the beer finishes very dry .  \textcolor{cyan}{medium   mouthfeel   and   medium   carbonation .} \\
    \bottomrule
    \end{tabular}
    \caption{A multi-aspect example from the BeerAdvocate dataset~\cite{beer}. \textcolor{blue}{Blue}, \textcolor{red}{red}, and \textcolor{cyan}{cyan} represent the aspects of \textit{Appearance}, \textit{Aroma}, and \textit{Palate}, respectively.}
    \label{table:beer_exp}
    % \vspace{-1em}
\end{table}

To address these problems, we propose the Multi-Aspect Rationale Extractor~(\textbf{MARE}). As shown in Figure~\ref{fig:MARE}, MARE can encode all aspects simultaneously by prepending multiple special tokens to the input text, each corresponding to a specific aspect. This approach enables multi-aspect encoding in one model. Furthermore, MARE introduces a Multi-Aspect Multi-Head Attention~(\textbf{MAMHA}) mechanism for collaborative encoding across aspects. This mechanism allows the model to capture interactions and dependencies between different aspects, leading to more accurate predictions and rationales. Finally, inspired by multi-task learning, MARE iteratively accesses training data for different aspects, reducing the overall training cost.

We validate the effectiveness of MARE on two unsupervised rationale extraction benchmarks: BeerAdvocate~\cite{beer} and Hotel Review~\cite{hotel}. Results show that MARE outperforms existing state-of-the-art methods across multiple evaluation metrics. Ablation studies further demonstrate the effectiveness of MARE.
Our main contributions are as follows:
\begin{itemize}
\item We introduce MARE, a Multi-Aspect Rationale Extractor that generates predictions and rationales for multiple aspects simultaneously.% MARE achieves an average improvement of XX\% over existing methods in terms of accuracy and efficiency. 
\item We deploy the multi-task training to reduce the training cost and expand the model applicability. Compared to multi-aspect collaborative training, it saves 17.9\% and 25.2\% of memory usage and training time, respectively.
\item Extensive experiments on BeerAdvocate and Hotel Review datasets demonstrate MARE's superiority, with a notable 5.4\% improvement in token-level F1 score. Ablation studies further validate the effectiveness of each component in MARE.
\end{itemize}

\section{Related Work}
\label{section:related_rationale}
The rationalization framework, known as RNP~\cite{lei-etal-2016-rationalizing}, assumes that any unselected input has no contribution to the prediction and achieves remarkable performance on this task. 
However, RNP still has many weaknesses. Various approaches have been proposed to improve RNP In different dimensions.

\paragraph{Gradient Flows} 
% REINFORCE~\cite{REINFORCE} is used in the base RNP framework, which introduces the instability of training and bad performance. 
% HardKuma~\cite{bastings-etal-2019-interpretable} introduces re-parameterization tricks and replaces the Bernoulii distribution to rectified Kumaraswamy distribution to stabilize the training process. 
% FR~\cite{FR} shares the encoder's parameter between the generator and predictor. The gradient of the encoder will be more reasonable because the encoder can see both full texts and rationales.
% 3Players~\cite{3player} controls the complementary rationale to be meaningless which makes the generated rationales more meaningful.
The RNP framework utilizes REINFORCE~\cite{REINFORCE} to overcome the non-differentiable problem, but this leads to training instability and poor performance.
HardKuma~\cite{bastings-etal-2019-interpretable} introduces re-parameterization tricks and replaces the Bernoulli distribution with the rectified Kumaraswamy distribution, which stabilizes the training process.
In FR~\cite{FR}, the encoder's parameter is shared between the generator and predictor. This ensures that the encoder's gradient is more reasonable because it can see both full texts and rationales.
3Players~\cite{3player} forces the complementary rationale to be meaningless, resulting in more meaningful generated rationales.
Our research is orthogonal with these methods.

\paragraph{Interlocking}
The interlocking problem was initially proposed by A2R~\cite{interlocking}. 
This problem arises when the generator fails to identify important tokens, leading to sub-optimal rationales and consequently affecting the performance.
Many researchers have developed approaches to address this issue~\cite{DMR, interlocking, MGR}.
DMR~\cite{DMR} aimed to align the distributions of rationales with the full input text in the output space and feature space.
A2R~\cite{interlocking} enhances the predictor's understanding of the full text by introducing a soft rationale.
MGR~\cite{MGR} involves multiple generators with different initializations to allow the predictor to see various rationales, alleviating the interlocking problem.
DR~\cite{DR} limits the Lipschitz constant of the predictor, making the whole system more robust.
DAR~\cite{DAR} deploys a pre-trained discriminator to align the selected rationale and the original input.
MCD~\cite{MCD} proposes the minimum conditional dependence criterion to overcome the issues of the maximum mutual information~(MMI) criterion.
YOFO~\cite{YOFO} eliminates interlocking by simultaneously predicting and interpreting. YOFO deploys pre-trained language models as its backbone and uses token deletion strategies between layers to erase unimportant tokens. the remaining tokens in the final layer are seen as rationales.

In the domain of multi-aspect rationale extraction, several approaches have been proposed. MTM~\cite{antognini2021multi} introduced a method for a multi-aspect explanation of target variables from documents, which bears some similarities to our work. Their approach, like ours, aims to provide explanations for multiple aspects simultaneously. However, there are key differences in the model architecture and methodology.
1. Model architecture: Unlike two-stage models that generate rationales and labels sequentially, MARE is a single-stage model that generates both simultaneously.
2. Base model: While some existing approaches use LSTM or CNN architectures, MARE leverages the power of pre-trained transformer models like BERT.
3. Aspect assignment: Our method allows for a token to be assigned to multiple aspects independently, whereas some existing methods normalize probabilities across aspects, limiting each token to a single aspect.

This paper focuses on the efficiency of the multi-aspect scenarios.
All the above models are uni-aspect encoding models, where one model can only encode one aspect of data.
MARE is a multi-aspect collaborative encoding model designed to encode multiple aspects of data simultaneously.

\begin{figure}
	\centering
	\subfloat[Attention Mask Deletion]{\includegraphics[width=0.4\linewidth]{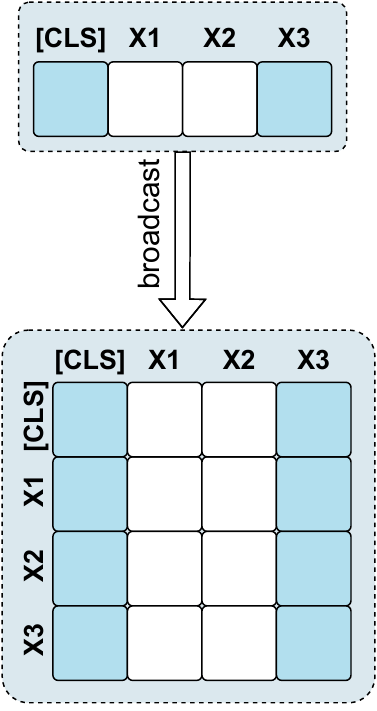}\label{fig:attention_matrix_broadcast}}
        \hspace{1em}
	\subfloat[Hard Deletion]{\includegraphics[width=0.4\linewidth]{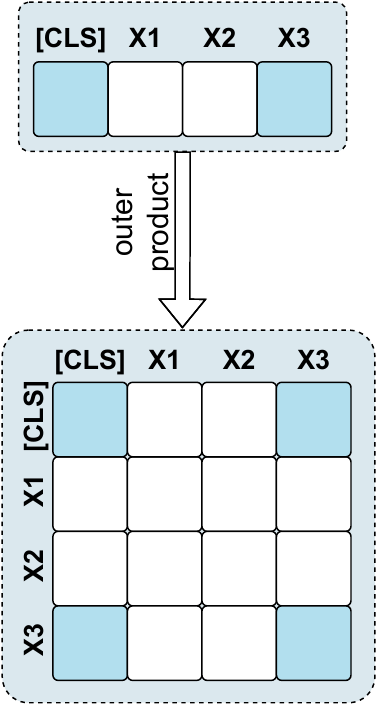}\label{fig:attention_matrix_exdot}} 
	\caption{Attention mask visualization. left: attention mask in Attention Mask Deletion. right: attention mask in Hard Deletion.
 }
    % \vspace{-1em}
\end{figure}

\section{Problem Definition}

Existing uni-aspect encoding models extract rationales $\mathbf{z}_i$ from the input $\mathbf{x}$ and predict the label $\mathbf{y}_i$ for the $i$-th aspect.
Formally, they can be expressed as $P(\mathbf{y}_i, \mathbf{z}_i\mid\mathbf{x};\theta_i)$, where $\theta_i$ represents the parameters of the model for the $i$-th aspect.
To obtain the rationales and predictions for all $k$ aspects, $k$ independently trained models are required: $\{P(\mathbf{y}_1, \mathbf{z}_1\mid\mathbf{x};\theta_1),\cdots, P(\mathbf{y}_k, \mathbf{z}_k\mid\mathbf{x};\theta_k)\}$.
However, this approach is time-consuming and computationally expensive.

To address this issue, we propose a multi-aspect rationale extraction task, where the rationales and predictions for all aspects can be generated simultaneously.
This can be formalized as $P(\mathbf{y}_1, \mathbf{z}_1, \cdots, \mathbf{y}_k, \mathbf{z}_k\mid\mathbf{x};\theta)$, where $\theta$ represents the parameters of the multi-aspect rationale extraction model.
By utilizing a single model to extract rationales and make predictions for all aspects concurrently, we aim to improve the efficiency and reduce computational costs compared.

%Previous uni-aspect encoding models extract rationales $\mathbf{z}_i$ from the input $\mathbf{x}$ and predict the label $\mathbf{y}_i$ for the $i$-th aspect.
%Formally, they can expressed as $P(\mathbf{y}_i, \mathbf{z}_i\mid\mathbf{x};\theta_i)$, where $\theta_i$ means parameters of the model of the $i$-th aspect.
%If we want to get the rationales and predictions for all $k$ aspects, then we need $k$ independently trained models, $\{P(\mathbf{y}_1, \mathbf{z}_1\mid\mathbf{x};\theta_1),P(\mathbf{y}_2, \mathbf{z}_2\mid\mathbf{x};\theta_2),\cdots, P(\mathbf{y}_k, \mathbf{z}_k\mid\mathbf{x};\theta_k)\}$.
%This is time-consuming and computationally expensive.

%To address the above issue, we propose the multi-aspect rationale extraction task where the rationales and predictions of all aspects can be generated simultaneously. 
%it can be formulized as $P(\mathbf{y}_1, \mathbf{z}_1, \cdots, \mathbf{y}_k, \mathbf{z}_k\mid\mathbf{x};\theta)$ where $\theta$ means the parameters of multi-aspect rationale extraction model.

\section{Method}
\begin{figure}[]
    \centering
    \includegraphics[width=0.48\textwidth]{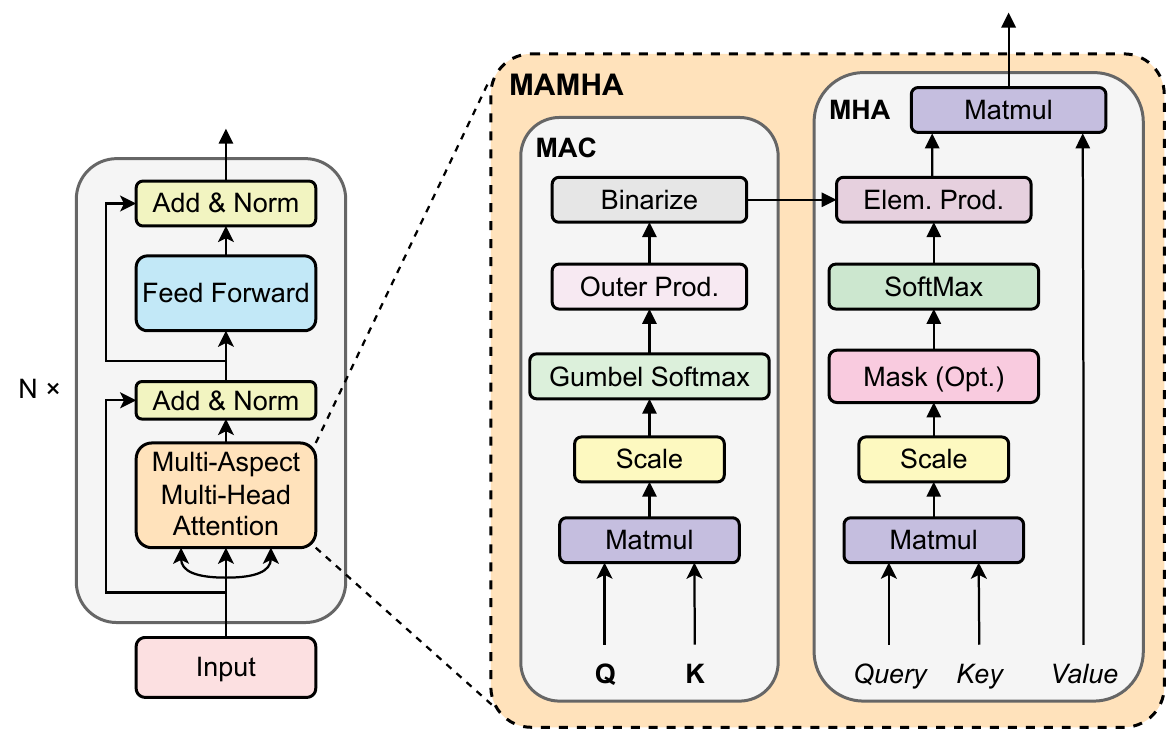}
    \caption{Overall model architecture. left: the overall model architecture of MARE. right: the computational graph of MAMHA. }
    \label{fig:mare_overall}
    % \vspace{-1em}
\end{figure}

% This paper proposes a Multi-Aspect Rationale Extractor, which can predict and interpret multiple aspects of text simultaneously.
% As shown in the left part of Figure~\ref {fig:mare_overall}, MARE is based on an encoder-based pre-trained language model and achieves multi-aspect collaborative encoding by a Multi-Aspect Multi-Head Attention mechanism. Meanwhile, MARE adopts multi-task training during the training process, greatly reducing the training cost.
This paper proposes a Multi-Aspect Rationale Extractor (MARE), which can simultaneously predict and interpret multiple aspects of text. As shown in the left part of Figure~\ref{fig:mare_overall}, MARE is based on an encoder-based pre-trained language model and achieves multi-aspect collaborative encoding through a Multi-Aspect Multi-Head Attention (MAMHA) mechanism. Additionally, MARE employs multi-task training during the training process, significantly reducing the training cost.
% These components of MARE will be introduced in the following sections.
% Section~\ref {sec:single_to_multiple} will provide a detailed introduction to the design concept and calculation method of MAMHA based on the "hard delete" operation.
% Section~\ref {sec:multi_task_training} will elaborate on multi-task training optimization for the training process.

\subsection{Hard Deletion for Complete Token Removal}
Selecting rationales without explicit annotations can be challenging. We follow the previous work~\cite{YOFO} where unimportant tokens are gradually erased. However, directly multiplying hidden states by the token mask harms rationalization performance~\cite{YOFO}. Attention Mask Deletion (AMD)~\cite{YOFO} avoids this problem by setting attention scores of masked tokens to 0. Concretely, assuming $\mathbf{m}_i\in[0, 1]^{L}$ represents the token mask in the $i$-th layer and $\mathbf{A}^j_i\in\mathbb{R}^{L\times L}$ is the attention score matrix of the $j$-th head in the $i$-th layer, the final attention score matrix is $\mathbf{\tilde{A}}^j_i=\mathbf{A}^j_i\cdot\mathbf{m}_i\in\mathbb{R}^{L\times L}$. Through AMD, remaining tokens interact while deleted ones are invisible.

However, AMD suffers from an "incomplete deletion" problem, where deleted tokens can still be partially represented by remaining ones due to the broadcast operation. As shown in Figure~\ref{fig:attention_matrix_broadcast}, although "X1" and "X2" are masked, they can still be indirectly represented by the weighted sum of "[CLS]" and "X3". Although this allows the model to retain more information, it hinders multi-aspect collaborative encoding. 

To address this issue, we propose \textbf{Hard Deletion}, which uses an outer product operation to completely erase deleted tokens (Figure~\ref{fig:attention_matrix_exdot}). "X1" and "X2" are represented by all-zero vectors, ensuring complete removal.

% \subsection{Multi-Aspect Multi-Head Attention}
% \label{sec:single_to_multiple}
% Inspired by the hard deletion, this paper proposes a new attention mechanism: multi-aspect multi-head attention mechanism, which can encode multiple text segments simultaneously. As shown in the right part of Figure~\ref{fig:mare_overall}, MAMHA consists of two parts: Multi Aspect Controller~(MAC) and the traditional multi-head attention~(MHA) mechanism.

\subsection{Multi-Aspect Multi-Head Attention}
\label{sec:single_to_multiple}
Inspired by hard deletion, we propose the multi-aspect multi-head attention (MAMHA) mechanism to encode multiple text segments simultaneously. As shown in the right part of Figure~\ref{fig:mare_overall}, MAMHA consists of a Multi-Aspect Controller (MAC) and the traditional multi-head attention (MHA) mechanism.

\begin{figure}[]
    \centering
    \includegraphics[width=0.48\textwidth]{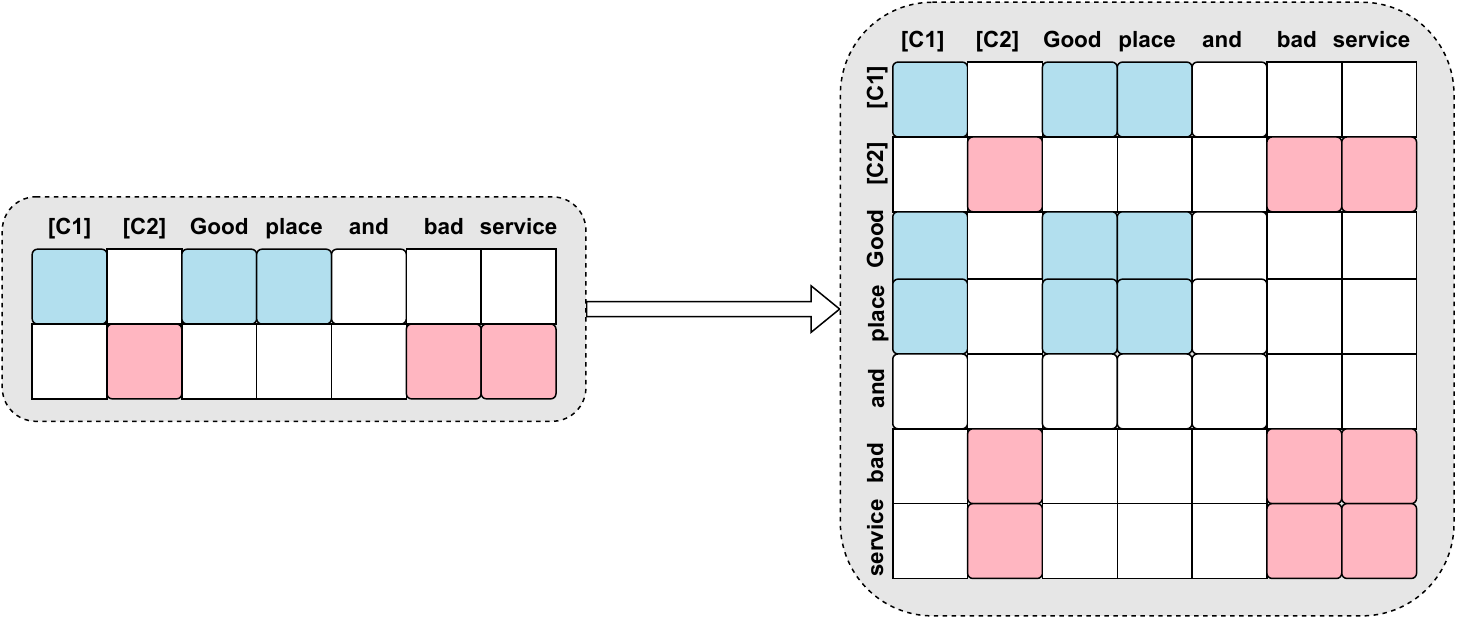}
    \caption{A example for Multi-Aspect Controller. left: The token mask for each aspect. "Good place" and "bad service" stands for the rationales of \textit{location} and \textit{service} aspect, respectively. right: The attention mask is obtained by performing an outer product operation on token masks.}
    \label{fig:attention_matrix_multiple}
    % \vspace{-1em}
\end{figure}

\subsubsection{Multi-Aspect Controller (MAC)}
MAC assists MHA in separately encoding different text segments by generating aspect-specific attention masks based on token masks for each aspect. This allows tokens within the same aspect to interact while isolating tokens from different aspects, enabling MHA to achieve multi-aspect collaborative encoding.

Figure~\ref{fig:attention_matrix_multiple} illustrates an example where "good place" and "bad service" are rationales for the "\textit{location}" and "\textit{service}" aspects, respectively. The final attention mask, obtained through an outer product operation, creates two separate segments. Words within each segment interact, while words from different segments remain independent. Special classification tokens "[C1]" and "[C2]" collect information from their respective aspects, allowing MHA to encode two aspects simultaneously.

This method can be extended to $k$ aspects by dividing the text into $k$ segments and appending $k$ special tokens. Note that if MAC employs AMD, tokens from different aspects cannot be fully isolated, leading to confusion and hindering multi-aspect collaborative encoding (further discussed in Section~\ref{sec:ablation_hard_soft}).

% \begin{figure}[]
%     \centering
%     \includegraphics[width=0.35\textwidth]{pic/mamha2.pdf}
%     \caption{Multi-Aspect Multi-Head Attention mechanism.}
%     \label{fig:mamha}
%     \vspace{-1em}
% \end{figure}

% \subsubsection{Computation Process}
% The computation process of MAC is shown in the right part of Figure~\ref {fig:mare_overall}.
% Specifically, assuming the hidden states of the $i$-th layer is $\mathbf{H}_i$, then its first $k$ vectors $\{\mathbf{h}_i^0,\cdots,\mathbf{h}_i^{k-1}\}$ are representations of special tokens.
% For the $j$-th aspect, the mapping functions $g_{query}^j$ and $g_{key}^j$ map special and normal tokens to $\mathbf{Q}$ and $\mathbf{K}$, respectively.
% Then calculate the similarity between special and normal tokens, and use the gumbel-softmax technique to determine whether a token is divided into a certain aspect.
% This process can be represented by Equation~(\ref{eq:tyofo_b})-(\ref{eq:tyofo_e}).
\subsubsection{Computation Process of MAC}
The computation process of MAC is shown in the right part of Figure~\ref{fig:mare_overall}. Assuming $\mathbf{H}_i$ represents the hidden states of the $i$-th layer, its first $k$ vectors $\{\mathbf{h}_i^0,\cdots,\mathbf{h}_i^{k-1}\}$ are representations of special tokens. For the $j$-th aspect, mapping functions $g_{query}^j$ and $g_{key}^j$ map special and normal tokens to $\mathbf{Q}$ and $\mathbf{K}$, respectively. The similarity between special and normal tokens is calculated, and the gumbel-softmax technique determines the token's aspect assignment (Equations~(\ref{eq:tyofo_b})-(\ref{eq:tyofo_e})).
\begin{gather}
    \label{eq:tyofo_b} \mathbf{Q} = \{g_{query}^0(\mathbf{h}_i^0), \cdots, g_{query}^k(\mathbf{h}_i^{k-1})\} \\
    \label{eq:tyofo_m} \mathbf{K} = \{g_{key}^0(\mathbf{H}_i[k:]), \cdots, g_{key}^k(\mathbf{H}_i[k:])\} \\
    \label{eq:tyofo_score} \mathbf{scores} = \frac{\mathbf{Q} \cdot \mathbf{K}^\mathsf{T}}{\sqrt{d}} \\
    \label{eq:tyofo_e} \mathbf{m} = \text{gumbel\_softmax}(\mathbf{scores}, dim=-1)
\end{gather}
, where $d$ and $L$ mean the vector's dimension and the text's length, respectively.
$[\cdot]$ represents slicing operation, $\mathbf{m}\in\{0,1\}^{k\times L}$ stands for the token mask, and $m[i, j]=1$ indicates that the $j$-th token is selected as the rationale of the $i$-th aspect.
% As shown in Equation~\ref{eq:ex_dot}, MAC adopts the outer product operation to match the shape of the attention score matrix in MHA.
The mask $\mathbf{m}$ evolves during training, starting as a near-full 1 vector and gradually becoming more selective. 
The Gumbel-Softmax output is binarized to produce the final mask.

MAC adopts the outer product operation to match the shape of the attention score matrix in MHA (Equation~\ref{eq:ex_dot}). When $M'[i, j]\neq 0$, the token is selected as a rationale in at least one aspect and should not be deleted (Equation~(\ref{eq:binarize})). The binarization operation in Equation~(\ref{eq:binarize}) is non-differentiable, so straight-through is used for gradient estimation. Finally, the mask is multiplied by the attention score matrix to perform token deletion (Equations~(\ref{eq:final_mask})-(\ref{eq:plm_nlayer})).
\begin{gather}
    \label{eq:ex_dot} \mathbf{M}'=\mathbf{m}^\mathsf{T}\cdot \mathbf{m} \in \mathbb{Z}\cap[0, k]^{L\times L} \\ 
    \label{eq:binarize} \tilde{M}[i, j]=\begin{cases}
        0\text{, If }M'[i, j]=0\\
        1\text{, Otherwise}
    \end{cases} \\
    \label{eq:final_mask} \mathbf{M}=\tilde{\mathbf{M}}+\mathbf{M}'-\text{StopGrad}(\mathbf{M}')\in[0, 1]^{L\times L} \\
    \label{eq:hard_mask_for_attention} \tilde{\mathbf{A}}_i^h=\mathbf{A}_i^h\odot\mathbf{M} \text{, for }h\text{ in }1,2,...,H \\
    \label{eq:plm_nlayer} \mathbf{H}_i=\mathrm{PLM}_i(\mathbf{H}_{i-1};\tilde{\mathbf{A}}_i) \text{, for }i\text{ in }1,2,...,N
\end{gather}
, where $\text{StopGrad}(X)$ represents stopping the X's gradient calculation.
$\mathbf{A}_i^h$ and $\tilde{\mathbf{A}}_i^h$ represent the initial and final attention score matrices of the $h$-th attention header in the $i$-th layer, respectively.
$\mathbf{H}_i$ represents the hidden layer representation of the $i$-th layer.

\subsection{Multi-Task Training}
\label{sec:multi_task_training}
Using labels from various aspects simultaneously during training may not be feasible, as datasets like Hotel Review~\cite{hotel} only have annotations for one aspect per sample. Multi-task training allows MARE to focus on the aspect corresponding to the current batch, avoiding the need to encode aspects with missing labels. If the batch comes from the $j$-th aspect, only the corresponding mapping functions $g_q^j$ and $g_k^j$ are used (Equations~(\ref{eq:train_yofov2_b})-(\ref{eq:train_yofov2_e})).
\begin{gather}
    \label{eq:train_yofov2_b} \mathbf{Q} = g_{query}^j(\mathbf{H}_i[j-1:j]) \\
    \label{eq:train_yofov2_e} \mathbf{K} = g_{key}^j(\mathbf{H}_i[k:])
\end{gather}
At inference time, we do not explicitly control the sparsity level. Instead, the trained mapping functions $g_q$ and $g_k$ directly select tokens they identify as explanations.
This means that the proportion of selected tokens for each aspect in a single sample is not strictly fixed. 
It can vary based on the content, and may even be 0\% if the model determines there is no relevant description for a particular aspect.

\subsection{Overall Loss}
\label{sec:overall_loss}
Our loss function consists of three components: cross-entropy loss ($L_{CE}$), sparsity penalty ($L_{sparse}$), and contiguous penalty ($L_{cont}$). The full loss function is:
\begin{gather}
    \label{eq:full_loss} L = L_{CE} + \beta L_{sparse} + \gamma L_{cont} \\
    L_{CE} = \frac1C\sum_{i=1}^{C}y_i\log{p_i} \\
    L_{sparse} = \frac1N\sum_{i=1}^N|\frac1L\sum_{j=1}^Lm_i^j-l_i| \\
    L_{cont} = \frac{\sum_{i=1}^N\sum_{j=1}^L|m_i^{j+1}-m_i^j|}{N(L-1)}
\end{gather}
, where $s$ is a predefined sparsity level, $\beta$ and $\gamma$ are hyperparameters that balance these terms.
We employ a Cliff decay strategy as illustrated in Appendix~\ref{appendix:cliff_decay}, where token deletion begins after a specified layer in the network.

\section{Experiments}
\subsection{Experimental Setup}
% \paragraph{BeerAdvocate} 
\paragraph{Datasets}
\label{sec:dataset2}
We performed experiments on two commonly used unsupervised rationale extraction datasets: BeerAdvocate~\cite{beer} and the Hotel Review dataset~\cite{hotel}.

The BeerAdvocate dataset~\cite{beer} is a multi-aspect sentiment prediction dataset. 
It consists of texts along with corresponding aspect scores ranging from 0 to 1, including aspects such as \textit{appearance}, \textit{aroma}, and \textit{palate}.
The training and validation sets do not have labeled rationales, but the test set contains 994 samples with rationale annotations for all aspects.
Notably, the scores across different aspects within the same sample exhibit high correlation, resulting in highly spurious correlations.
For the BeerAdvocate dataset, we conducted experiments on the decorrelated version proposed by \citeauthor{lei-etal-2016-rationalizing}.
We binarized the dataset into binary classification tasks using a positive threshold of 0.6 and a negative threshold of 0.4~\cite{bao2018deriving}.
We run our model, MARE, on two sparsity levels: high-sparse and low-sparse.
In the high-sparse decorrelated dataset, the sparsity level approximates the sparsity for golden rationales in the test set.
In the low-sparse decorrelated dataset, the sparsity level is comparatively lower but allows for convenient comparisons with previous works.
% To examine the susceptibility of our model to spurious correlations, we also utilized the correlated BeerAdvocate dataset by \citeauthor{MGR}.

\begin{table*}[]
    \centering
    
    \resizebox{\textwidth}{!}
    {
    \begin{tabular}{l|lllll|lllll|lllll|l}
    \toprule
    \multirow{2}{*}{Methods} & \multicolumn{5}{c|}{Appearance}                                                                                           & \multicolumn{5}{c|}{Aroma}                                                                                                & \multicolumn{5}{c|}{Palate}         & Avg                                                                                      \\\cline{2-16} 
                             & \multicolumn{1}{c}{S} & \multicolumn{1}{c}{ACC} & \multicolumn{1}{c}{P} & \multicolumn{1}{c}{R} & \multicolumn{1}{c|}{F1} & \multicolumn{1}{c}{S} & \multicolumn{1}{c}{ACC} & \multicolumn{1}{c}{P} & \multicolumn{1}{c}{R} & \multicolumn{1}{c|}{F1} & \multicolumn{1}{c}{S} & \multicolumn{1}{c}{ACC} & \multicolumn{1}{c}{P} & \multicolumn{1}{c}{R} & \multicolumn{1}{c|}{F1} & \multicolumn{1}{c}{F1} \\\hline
    RNP\cite{lei-etal-2016-rationalizing}                     & 18.7                  & 84.0                    & 72.0                  & 72.7                  & 72.3                    & 15.1                  & 85.2                    & 59.0                  & 57.2                  & 58.1                    & 13.4                  & 90.0                    & 63.1                  & 68.2                  & 65.5              &   65.6  \\
    DMR\cite{DMR}                     & 18.2                  & -                       & 71.1                  & 70.2                  & 70.7                    & 15.4                  & -                       & 59.8                  & 58.9                  & 59.3                    & 11.9                  & -                       & 53.2                  & 50.9                  & 52.0 &       60.7           \\
    A2R\cite{interlocking}                     & 18.4                  & 83.9                    & 72.7                  & 72.3                  & 72.5                    & 15.4                  & 86.3                    & 63.6                  & 62.9                  & 63.2                    & 12.4                  & 81.2                    & 57.4                  & 57.3                  & 57.4         &      64.5    \\
    FR\cite{FR}                     & 18.4                  & {87.2}              & 82.9                  & 82.6                  & 82.8                    & 15.0                  & {88.6}           & 74.7                  & 72.1                  & 73.4                    & 12.1                  & {89.7}           & 67.8                  & 66.2                  & 67.0               & 74.4   \\
    MGR\cite{MGR}                     & 18.4                  & 86.1                    & 83.9                  & 83.5                  & 83.7                    & 15.6                  & 86.6                    & 76.6                  & 76.5                  & 76.5                    & 12.4                  & 85.1                    & 66.6                  & 66.6                  & 66.6 &        75.6          \\
    DR\cite{DR}                      & 18.6                  & 85.3                    & 84.3                  & 84.8                  & 84.5                    & 15.6                  & {87.2}              & 77.2                  & 77.5                  & 77.3                    & 13.3                  & 85.7                    & 65.1                  & 69.8                  & 67.4 &       76.4 \\%\hline
    % BERT-RNP                 & 18.8                  & {86.4}                    & 83.2                  & 82.3                  & 82.8                    & 15.2                  & 82.8                    & 77.4                  & 70.8                  & 73.9                    & 13.0                  & 87.3                    & 64.5                  & 63.1                  & 63.8                   \\
    % BERT-FR                  & 21.3                  & 84.2                    & 24.2                  & 27.2                  & 25.6                    & 17.9                  & 83.8                    & 51.6                  & 55.8                  & 53.6                    & 12.2                  & 88.0                    & 57.2                  & 52.6                  & 54.8                   \\\hline
    % YOFO-final               & 18.4                  & \textbf{87.4}           & 90.9                  & {88.0}         & {89.5}                    & 14.7                  & 85.7                    & 92.9                  & 82.5                  & 87.4                    & 13.6                  & 87.8                    & 72.2                  & 73.5                  & 72.8                   \\
    YOFO~\cite{YOFO}                     & 18.1                  & 85.6                    & {91.3}                  & {87.1}                  & 89.2                    & 15.4                  & 86.8                    & \textbf{94.3}                  & {87.9}         & {91.0}           & 13.2                  & {88.4}              & {79.5}                  & {79.0}            & {79.2}       &   86.5        \\%\hline
    % multi-YOFO               & 18.4                  & 85.8                    & \textbf{97.0}         & {87.5}            & \textbf{92.0}           & 15.5                  & 85.1                    & \textbf{97.2}         & 84.4                  & {90.3}              & 12.6                  & 87.2                    & {84.0}            & \textbf{80.1}         & \textbf{82.0}          \\
    MARE~(\textit{ours})                    & 17.3                  & 85.6                    & \textbf{95.4}            & \textbf{89.7}                  & \textbf{92.5}              & 15.4                  & 86.0                    & {93.9}            & \textbf{90.2}            & \textbf{92.0}                    & 12.7                  & 88.0                    & \textbf{82.2}         & \textbf{81.9}                  & \textbf{82.0}    & \textbf{88.8}   \\\bottomrule     
    \end{tabular}
    }
    \caption{Results of different methods on the high-sparse decorrelated BeerAdvocate dataset. %“*”和“**”表示从获得的结果\cite{dr}and~\cite{MGR}。\textbf{粗体}表示不同方法中的最佳结果。\underline{下划线}表示不同方法中的第二名结果。
    }
    \label{table:high_sparse_beer2}
\end{table*}

\begin{table*}[]
    \centering
    
    \resizebox{\textwidth}{!}
    {
    \begin{tabular}{l|lllll|lllll|lllll|l}
    \toprule
    \multirow{2}{*}{Methods} & \multicolumn{5}{c|}{Appearance}                                                                                                                           & \multicolumn{5}{c|}{Aroma}                                                                                                                                & \multicolumn{5}{c|}{Palate}    & Avg                                                                                                                            \\\cline{2-16} 
                             & \multicolumn{1}{c}{S} & \multicolumn{1}{c}{ACC}        & \multicolumn{1}{c}{P}          & \multicolumn{1}{c}{R}          & \multicolumn{1}{c|}{F1}        & \multicolumn{1}{c}{S} & \multicolumn{1}{c}{ACC}        & \multicolumn{1}{c}{P}          & \multicolumn{1}{c}{R}          & \multicolumn{1}{c|}{F1}        & \multicolumn{1}{c}{S} & \multicolumn{1}{c}{ACC}        & \multicolumn{1}{c}{P}          & \multicolumn{1}{c}{R}          & \multicolumn{1}{c|}{F1}   & \multicolumn{1}{c}{F1}      \\\hline
    RNP\cite{lei-etal-2016-rationalizing}                     & 11.9                  & -                              & 72.0                           & 46.1                           & 56.2                           & 10.7                  & -                              & 70.5                           & 48.3                           & 57.3                           & 10.0                  & -                              & 53.1                           & 42.8                           & 47.5                   &    53.7    \\
    CAR\cite{CAR}                     & 11.9                  & -                              & 76.2                           & 49.3                           & 59.9                           & 10.3                  & -                              & 50.3                           & 33.3                           & 40.1                           & 10.2                  & -                              & 56.6                           & 46.2                           & 50.9             &     50.3         \\
    DMR\cite{DMR}                     & 11.7                  & -                              & 83.6                           & 52.8                           & 64.7                           & 11.7                  & -                              & 63.1                           & 47.6                           & 54.3                           & 10.7                  & -                              & 55.8                           & 48.1                           & 51.7              &       56.9      \\
    FR\cite{FR}                     & 12.7                  & 83.9                           & 77.6                           & 53.3                           & 63.2                           & 10.8                  & {87.6} & 82.9                           & 57.9                           & 68.2                           & 10.0                  & 84.5                           & 69.3                           & 55.8                           & 61.8                       & 64.4   \\
    MGR\cite{MGR}                      & 13.2                  & 82.6                           & 75.2                           & 53.5                           & 62.6                           & 12.3                  & 84.7                       & 80.8                           & 63.7                           & 71.2                           & 10.8                  & 80.1                           & 51.6                           & 44.7                           & 47.9            &      60.6         \\
    DR\cite{DR}                      & 11.9                  & 81.4                           & 86.8                           & 55.9                           & 68.0                           & 11.2                  & 80.5                           & 70.8                           & 57.1                           & 63.2                           & 10.5                  & 81.4                           & 71.2                           & 60.2                           & 65.3               &     65.5       \\%\hline
    % BERT-RNP                 & 13.8                  & 84.9                           & 84.7                           & 61.4                           & 71.2                           & 13.9                  & 86.2                           & 70.9                           & 59.5                           & 64.7                           & 11.9                  & {85.1}                           & 73.1                           & 65.5                           & 69.1                           \\
    % BERT-FR                  & 15.4                  & 84.4                           & 22.8                           & 18.5                           & 20.4                           & 14.4                  & 85.5                           & 21.4                           & 18.7                           & 19.9                           & 14.1                  & 86.6                           & 16.6                           & 17.5                           & 17.1                           \\
    % YOFO-final               & 13.2                  & \textbf{87.5} & {97.4}                           & {67.6}                           & {79.8}                           & 12.4                  & 85.5                           & {95.4}                           & {71.2}                           & {81.6}                           & 10.2                  & 87.6                           & 83.3                           & 63.7                           & 72.2                           \\
    YOFO~\cite{YOFO}                     & 13.1                  & {87.0}                           & {97.1}                           & {66.9}                           & {79.2}                           & 12.1                  & {86.3}                           & {94.1}                           & {68.9}                           & {79.5}                           & 10.9                  & {87.8} & \textbf{88.5}                           & {72.7} & {79.8}  & 79.5 \\%\hline
    % multi-YOFO               & 13.2                  & \textbf{87.1}     & {98.4}     & \textbf{68.3} & \textbf{80.6} & 13.4                  & {86.6}     & \textbf{97.7} & {71.5}     & \textbf{82.6} & 11.3                  & {87.7}     & \textbf{90.3} & \textbf{72.7} & \textbf{80.5} \\
    MARE~(\textit{ours})                    & 13.8                  & 86.3                           & \textbf{98.7} & \textbf{74.0}     & \textbf{84.6}     & 12.2                  & 85.9                           & \textbf{97.5}     & \textbf{74.4} & \textbf{84.3} & 10.9                  & {88.2} & {87.4}     & \textbf{74.6}     & \textbf{80.5}                  &    \textbf{83.1}     \\\bottomrule
    \end{tabular}
    }
    \caption{Results of different methods on the low-sparse decorrelated BeerAdvocate dataset.%“*”和“**”表示从获得的结果\cite{dr}and~\cite{MGR}。\textbf{粗体}表示不同方法中的最佳结果。\underline{下划线}表示不同方法中的第二名结果。
    }
    \label{table:low_sparse_beer2}
    % \vspace{-1em}
\end{table*}

The Hotel Review dataset~\cite{hotel} is another widely used dataset for multi-aspect sentiment classification and rationale extraction. It includes texts along with three aspect labels: \textit{location}, \textit{service}, and \textit{cleanliness}. 
In addition to the aspect labels, the test set of this dataset also provides rationale annotations for all three aspects, with 200 samples. Since the original labels are on a scale of 0 to 5 stars, we utilize the binarized version proposed by \citeauthor{bao2018deriving}.
For the Hotel Review dataset, we only conducted a low-sparse experiment as the golden sparsity level is relatively low, at around 10\%.

The statistics of the BeerAdvocate~\cite{beer} and Hotel Review dataset~\cite{hotel} are shown in Table~\ref{tab:statistics}.
\begin{table}[!htpb]
\centering
\small
\resizebox{0.5\textwidth}{!}{
\begin{tabular}{cc|cc|cc|cc}
\toprule
\multicolumn{2}{c|}{\multirow{2}{*}{Datasets}} & \multicolumn{2}{c|}{Train} & \multicolumn{2}{c|}{Validation} & \multicolumn{2}{c}{Test} \\
\multicolumn{2}{c|}{}                          & Pos          & Neg         & Pos            & Neg            & Pos         & Neg        \\ \midrule
\multirow{3}{*}{Beer}       & Appearance       & 16891        & 16891       & 6628           & 2103           & 923         & 13         \\
                            & Aroma            & 15169        & 15169       & 6579           & 2218           & 848         & 29         \\
                            & Palate           & 13652        & 13652       & 6740           & 2000           & 785         & 20         \\ \hline
\multirow{3}{*}{Hotel}      & Location         & 7236         & 7236        & 906            & 906            & 104         & 96         \\
                            & Service          & 50742        & 50742       & 6344           & 6344           & 101         & 98         \\
                            & Cleanliness      & 75049        & 75049       & 9382           & 9382           & 97          & 99  \\
                            \bottomrule
\end{tabular}
}
\caption{Statistics of the BeerAdvocate and Hotel Review dataset.}
\label{tab:statistics}
\end{table}

\paragraph{Baselines}
\label{sec:baseline2}
We compared the performance of MARE with several state-of-the-art baselines. 
These baselines, including RNP~\cite{lei-etal-2016-rationalizing}, CAR~\cite{CAR}, DMR~\cite{DMR}, A2R~\cite{interlocking}, FR~\cite{FR}, MGR~\cite{MGR} DR~\cite{DR}, and YOFO~\cite{YOFO}, were discussed in Section~\ref{section:related_rationale}.
The performance of these baselines are obtained from YOFO~\cite{YOFO}.
In MARE, we use BERT for our backbone and the balanced round-robin is equipped in the training stage.
All of our experiments are conducted on NVIDIA Geforce RTX 3090 24GB. 
For more implementation details, please refer to Appendix~\ref{sec:setting2}.
% Our methods, YOFO with the default setting (YOFO) and YOFO that only constrains sparsity in the final layer (YOFO-final), were used for comparison. 
% To ensure a fair comparison, we implemented the following baselines since we utilized a pre-trained language model in YOFO.
% BERT-RNP is the RNP framework equipped with BERT, which serves as our direct baseline. 
% BERT-FR is similar to BERT-RNP, except that the parameters of the generator and predictor are shared.
% We also re-implemented most of the baselines mentioned above, and the results can be found in Appendix~\ref{sec:reimp}.

\paragraph{Metrics}
\label{sec:metrics2}
Following previous works~\cite{YOFO}, we will use token-level F1 and accuracy for the rationalization and downstream performance.
In our result tables, we define S as the sparsity level of selected rationales, computed using the formula $S=\frac{\#selected\text{ }tokens}{\#tokens}$. 
P, R, and F1 represent precision, recall, and F1 score for rationale extraction, respectively. 
ACC and Val ACC denote the accuracy of the test and validation sets, respectively. 
The best performance is \textbf{Bolded} in the tables.

\subsection{Main Results}
\subsubsection{Results on the BeerAdvocate Dataset}
\paragraph{High-sparse}
Experimental results on the de-correlated BeerAdvocate dataset in the high-sparse scenario are shown in Table~\ref{table:high_sparse_beer2}.
MARE outperforms YOFO by 3.3\%, 1.0\%, and 2.8\% in the \textit{appearance}, \textit{aroma}, and \textit{palate} aspects, respectively.
Meanwhile, MARE achieves the best average F1 scores among all models, particularly 88.8\%.
% This indicates that MARE outperforms YOFO while reducing the number of parameters and computations.
This is because MARE is a multi-aspect collaborative encoding model that captures internal correlations between all aspects and thus achieves the best performance.

% For detailed case analysis, please refer to section \ref{sec:caseubeer2}.
% The performance of YOFO and MARE in the rationale extraction task is far superior to other baseline models, indicating that the single-stage rationale extraction model is extremely powerful.
% However, in terms of accuracy, the YOFO series models did not perform as well as expected.
% The reason is that the model was trained on the training set for 10 to 15 epochs, which resulted in severe overfitting of the YOFO series models.
% Meanwhile, on the BeerAdvocate dataset, the sample distribution on the test set is extremely uneven (with a positive-to-negative sample ratio of approximately 41:1), which to some extent affects the performance of the model.

\begin{table*}[]
    \centering

    \resizebox{\textwidth}{!}
    {
    \begin{tabular}{l|lllll|lllll|lllll|l}
    \toprule
    {\multirow{2}{*}{Methods}} & \multicolumn{5}{c|}{Location}                                                                                             & \multicolumn{5}{c|}{Service}                                                                                              & \multicolumn{5}{c|}{Cleanliness}   & Avg                                                                                       \\\cline{2-16} 
    \multicolumn{1}{c|}{}                         & \multicolumn{1}{c}{S} & \multicolumn{1}{c}{ACC} & \multicolumn{1}{c}{P} & \multicolumn{1}{c}{R} & \multicolumn{1}{c|}{F1} & \multicolumn{1}{c}{S} & \multicolumn{1}{c}{ACC} & \multicolumn{1}{c}{P} & \multicolumn{1}{c}{R} & \multicolumn{1}{c|}{F1} & \multicolumn{1}{c}{S} & \multicolumn{1}{c}{ACC} & \multicolumn{1}{c}{P} & \multicolumn{1}{c}{R} & \multicolumn{1}{c|}{F1} & \multicolumn{1}{c}{F1} \\\hline
    RNP\cite{lei-etal-2016-rationalizing}                                         & 8.8                   & {97.5}                    & 46.2                  & 48.2                  & 47.1                    & 11.0                  & {97.5}                    & 34.2                  & 32.9                  & 33.5                    & 10.5                  & 96.0                    & 29.1                  & 34.6                  & 31.6           &    37.4    \\
    DMR\cite{DMR}                                          & 10.7                  & -                       & 47.5                  & 60.1                  & 53.1                    & 11.6                  & -                       & 43.0                  & 43.6                  & 43.3                    & 10.3                  & -                       & 31.4                  & 36.4                  & 33.7           &    43.4    \\
    A2R\cite{interlocking}                                          & 8.5                   & 87.5                    & 43.1                  & 43.2                  & 43.1                    & 11.4                  & 96.5                    & 37.3                  & 37.2                  & 37.2                    & 8.9                   & 94.5                    & 33.2                  & 33.3                  & 33.3       &   37.9         \\
    FR\cite{FR}                                           & 9.0                   & 93.5                    & 55.5                  & 58.9                  & 57.1                    & 11.5                  & 94.5                    & 44.8                  & 44.7                  & 44.8                    & 11.0                  & 96.0                    & 34.9                  & 43.4                  & 38.7          &       46.9  \\
    MGR\cite{MGR}                                         & 9.7                   & {97.5}                    & 52.5                  & 60.5                  & 56.2                    & 11.8                  & 96.5                    & 45.0                  & 46.4                  & 45.7                    & 10.5                  & 96.5                    & 37.6                  & 44.5                  & 40.7      &   47.5         \\
    DR\cite{DR}                                           & 9.6                   & 96.5                    & 53.6                  & 60.9                  & 57.0                    & 11.5                  & 96.0                    & 47.1                  & 47.4                  & 47.2                    & 10.0                  & {97.0}                    & 39.3                  & 44.3                  & 41.8         &   48.9       \\ % \hline
    % BERT-RNP                                      & 9.4                   & \textbf{98.0}              & 47.2                  & 49.9                  & 48.5                    & 10.2                  & \textbf{99.5}           & 39.1                  & 33.1                  & 35.8                    & 9.8                   & \textbf{100.0}          & 45.1                  & 45.2                  & 45.2                   \\
    % BERT-FR                                       & 8.5                   & \textbf{98.5}           & 14.3                  & 13.9                  & 14.1                    & 10.8                  & \textbf{99.5}           & 26.4                  & 23.8                  & 25.0                    & 11.1                  & \textbf{100.0}          & 25.6                  & 29.4                  & 27.4                   \\
    % YOFO-final                                    & 10.9                  & \textbf{98.0}              & 54.2                  & {64.1}         & {58.8}                    & 10.5                  & \textbf{99.5}           & \textbf{60.8}         & {55.9}                  & \textbf{58.3}           & 9.3                   & \textbf{100.0}          & {47.5}                  & 45.2                  & 46.3                   \\
    YOFO~\cite{YOFO}                                          & 9.7                   & {98.0}              & {55.7}                  & 60.4                  & 58.0                    & 11.9                  & {99.5}           & {58.3}            & \textbf{57.4}                  & \textbf{57.9}              & 10.6                  & {100.0}          & \textbf{49.9}                  & \textbf{54.4}         & \textbf{52.1}       &  56.0 \\%\hline
    % multi-YOFO                                    & 11.4                  & \textbf{98.0}              & {58.9}            & {62.3}                  & {60.5}              & 11.1                  & \textbf{99.5}           & 52.4                  & {61.2}         & {56.5}                    & 14.7                  & {99.5}              & {54.4}            & 44.7                  & 49.1                   \\
    MARE~(\textit{ours})                                         & 9.7                  & {98.0}              & \textbf{59.0}         & \textbf{68.4}            & \textbf{63.3}           & 10.8                  & {99.5}              & \textbf{58.6}                  & {55.2}            & 56.8                    & 10.6                  & {100.0}              & {46.8}         & {54.0}            & {50.1}          &   \textbf{56.7} \\\bottomrule
    \end{tabular}
    }
    \caption{Results of different methods on the Hotel Review dataset. %“*”和“**”表示从获得的结果\cite{dr}and~\cite{MGR}。\textbf{粗体}表示不同方法中的最佳结果。\underline{下划线}表示不同方法中的第二名结果。
    }
    \label{table:hotel2}
\end{table*}

\begin{table*}
    \centering
    \small
    \resizebox{\textwidth}{!}{
    \begin{tabular}{p{15cm}} 
    \toprule
    \hline
    MARE~(\textit{ours})\\
    \hline
    \textbf{Location}: \textcolor{blue}{Positive} \ding{52} \\ 
    \textbf{Service}: \textcolor{red}{Positive} ? \\
    \textbf{Cleanliness}: \textcolor{cyan}{Positive} ? \\
    \textbf{Text}: arrived very apprehensively to the hotel after reading the negative remarks . we were happily suprised \textcolor{red}{. staff very pleasant ,} \textcolor{cyan}{rooms and bathrooms spotlessly clean , although on the small} side . our rooms had no natural light , but with the lights on were ok . air conditioning worked ( was necessary in november ! ) , although noisy from the inside and outside where the vents are . however \textcolor{blue}{,} the hotel \textcolor{blue}{is in the middle of nyc and} the noise did n ' t bother us overmuch - the situation is much more important , \textcolor{blue}{and the jolly was in} \textcolor{blue}{\uline{a perfect location}}\uline{ for shopping and tourism} . breakfasted across the road in the moonstruck deli ( opens at 7 am for the jet lagged ) . i would certainly go back there again ! \\
    \hline
    \hline
    \textbf{Location}: -  \\ 
    \textbf{Service}: \textcolor{red}{Positive} \ding{52} \\
    \textbf{Cleanliness}: - \\
    \textbf{Text}: this is a very nice hotel \textcolor{red}{with \uline{top - notch service and staff} .} you will pay for it , but if you want to avoid the touristy hotels of branson , this is a beautiful place to stay and eat .  \\
    \hline
    \bottomrule
    \end{tabular}
    }
    \caption{Case studies on the Hotel Review dataset. %Golden rationales are marked using \ul{underline}. Our rationales are highlighted using \textcolor{cyan}{cyan}.}
    }
    \label{table:decode_hotel2}
    % \vspace{-1em}
\end{table*}

\paragraph{Low-sparse}
Experimental results on the de-correlated BeerAdvocate dataset in the low-sparse scenario are shown in Table~\ref{table:low_sparse_beer2}.
% The performance of MARE was also measured in experiments on the BeerAdvocate dataset in low sparsity scenarios, as shown in Table \ref {table:low_sparse_beer2}.
MARE still achieves the best performance in all aspects, similar to the high-sparsity scenario.
In the low-sparsity scenario, the performance gain obtained by MARE is greater than in high-sparsity scenarios.
Specifically, MARE is 5.4\%, 4.8\%, and 0.7\% higher than YOFO in the \textit{appearance}, \textit{aroma}, and \textit{palate} aspects, respectively.
Furthermore, MARE has a 3.6\% average performance gain in token-level F1 compared to YOFO.
This further demonstrates the effectiveness of MARE.

\subsubsection{Results on the Hotel Review Dataset}
Experimental results on the Hotel Review dataset are shown in Table~\ref{table:low_sparse_beer2}.
% The effectiveness of MARE was validated through experiments on the Hotel Review dataset, and the results are shown in Table \ref {table:notel2}.
% The conclusions are significantly different from those on the BeerAdvocate dataset.
Although MARE is slightly inferior to YOFO in the \textit {service} and \textit {cleanliness} aspects, it is far superior to YOFO in the \textit{location} aspect and its average token-level F1 score is higher than YOFO.
Specifically, MARE is 1.1\% and 2.0\% lower than YOFO in the \textit {service} and \textit {cleanliness} aspects, respectively, while it is 5.3\% higher than YOFO in the \textit{location} aspect.
Meanwhile, MARE is 0.7\% higher than YOFO in the average token-level F1 score.

\begin{table*}[]
    \centering
    \small
    \resizebox{\textwidth}{!}{
    \begin{tabular}{ccccccccc}
    \toprule
    \multicolumn{1}{c}{\multirow{2}{*}{Methods}} & \multicolumn{1}{c}{\multirow{2}{*}{\begin{tabular}[c]{@{}c@{}}Memory Usage\\(MB)\end{tabular}}} & \multicolumn{1}{c}{\multirow{2}{*}{\begin{tabular}[c]{@{}c@{}}Training Time\\(minutes/epoch)\end{tabular}}} & \multicolumn{2}{c}{Appearance}                     & \multicolumn{2}{c}{Aroma}                     & \multicolumn{2}{c}{Palate}                     \\\cline{4-9} 
    \multicolumn{1}{c}{}                    & \multicolumn{1}{c}{}                                                                     & \multicolumn{1}{c}{}                                                                       & \multicolumn{1}{c}{ValAcc} & F1            & \multicolumn{1}{c}{ValAcc} & F1            & \multicolumn{1}{c}{ValAcc} & F1            \\\midrule
    multi-aspect collaborative training                                 & 24209                                                                                    & 34.5                                                                                       & \textbf{89.2}              & 92.2          & 88.4                       & 90.1          & 84.0                       & 79.2          \\
    multi-task training                                 & \textbf{19877}                                                                           & \textbf{25.8}                                                                              & \textbf{89.2}              & \textbf{92.5} & \textbf{89.1}              & \textbf{92.0} & \textbf{84.7}              & \textbf{82.0} \\\bottomrule
    \end{tabular}
    }
    \caption{Ablation study on different training strategies.}
    \label{table:strategy}
    % \vspace{-1em}
\end{table*}

\section{Analysis}
In this section, we delve into a more comprehensive analysis of our methodology.
In Section~\ref{sec:case}, we present a series of case studies derived from the Hotel Review dataset to exemplify the practical applications of our approach.
In Section~\ref{sec:ablation}, we conduct an ablation study to substantiate the efficacy of our method by incrementally removing its constituent elements.
\subsection{Case Study}
\label{sec:case}
This section visualizes several samples on the Hotel Review dataset as shown in Table~\ref{table:decode_hotel2}.
\textcolor{blue}{Blue},  \textcolor{red}{red}, and \textcolor{cyan}{cyan} represent the \textit {location}, \textit {service}, and \textit {cleanliness} aspects, respectively, and \uline{underline} indicate the annotated rationales.

In the Hotel Review test set, each sample only has a uni-aspect annotation.
As shown in the first case, only the \textit{location} aspect has been annotated.
However, in real scenarios, a review often describes multiple aspects.
MARE extracted snippets not only about \textit{location} but \textit {service} and \textit {cleanliness} which are not annotated.
"\textit{Staff very clean}" and "\textit{rooms and bathrooms spotless clean}" demonstrate that the \textit {service} and \textit {cleanliness} of the hotel are excellent.
In the second case, only the \textit {location} aspect appeared in the text.
Correspondingly, MARE did not select any rationale other than the \textit {location} aspect.
This indicates that MARE benefits from multi-aspect collaborative encoding and makes decisions when there is clear evidence.

\subsection{Ablation Studies}
\label{sec:ablation}
To verify the effectiveness of our model components, we have conducted several ablation studies on the BeerAdvocate dataset~\cite{beer}.

\subsubsection{multi-task training v.s. multi-aspect collaborative training}
\label{sec:ablation_multitask_multiaspect}
To explore the impact of multi-task training on the model as described in Section~\ref{sec:multi_task_training}, this experiment verifies the effectiveness of multi-task training by comparing the performance, memory usage, and time cost of multi-task training and multi-aspect collaborative training.

% MARE is trained on NVIDIA RTX 3090 24GB. 
The experimental result is shown in Table~\ref{table:strategy}.
The performance of multi-task training is slightly better than that of multi-aspect collaborative training. 
% The performance of the two is similar in the "appearance" aspect, but better performance is achieved in the more difficult "aroma" and "taste" aspects through multitasking training. 
This is because, in the early stages of training, MARE cannot distinguish various aspects well, so multi-aspect collaborative training may lead to information leakage between different aspects, resulting in a performance drop. 
% Although this issue gradually improves with training, its performance is still not as good as the model trained through multitasking. 
Meanwhile, multi-aspect collaborative training requires mask calculation for all aspects, resulting in high memory usage and long training time, reaching 24209MB and 34.5 minutes respectively. 
By contrast, multi-task training only requires encoding a single aspect at a time, so it costs much lower in both memory and training time. 
It saves 17.9\% and 25.2\% of memory usage and training time, respectively.
This indicates that models trained using multi-task training can outperform those trained using multi-aspect collaborative training with fewer computational resources, demonstrating the effectiveness of multi-task training.

\subsubsection{Hard Deletion v.s. Attention Mask Deletion}
\label{sec:ablation_hard_soft}
To demonstrate the effectiveness of hard deletion, this section contrastively employs AMD operations in the MAC.
% Specifically, unlike the calculation method in Equation~(\ref{eq:ex_dot}), the mask $\mathbf{m}$ obtains a mask vector $\mathbf{m'}$ through an addition operation, which is then multiplied by the attention score matrix through the broadcast mechanism. 
Specifically, we will replace the Equation~(\ref {eq:ex_dot})-(\ref{eq:hard_mask_for_attention}) with Equation~(\ref{eq:soft_deletion_ablation_start})-(\ref{eq:soft_deletion_ablation_end}):
\begin{gather}
\label{eq:soft_deletion_ablation_start} \mathbf{m'} = \sum_{i=0}^{k-1}\mathbf{m}[i]\in [0,k]^{L} \\
    \tilde{m}[i]=\begin{cases}
        0\text{, If } m'[i]=0\\
        1\text{, Otherwise}
    \end{cases} \\
    \hspace{1.3em}\hat{\mathbf{m}} = \mathbf{m'} - \text{StopGrad}(\mathbf{m'}) + \tilde{\mathbf{m}}\in \{0,1\}^{L} \\
    \label{eq:soft_deletion_ablation_end} \tilde{\mathbf{A}}_i^h=\mathbf{A}_i^h\odot\hat{\mathbf{m}} \text{, for }h\text{ in }1,2,...,H
\end{gather}
, where $k$ means the number of aspects, and $\mathbf{m'}$ represents the mask vector with a span of closed interval [0, k], $\hat{\mathbf{m}}$ indicates the calculated mask vector to multiply with attention score matrix.
Here, we also use the Straight Through technique to bypass the non-differentiable problem. 

Experimental results are shown in Table~\ref{table:deletion}. 
While using AMD, the rationalization and downstream performance are very poor. 
On the contrary, MARE-hard performs very well. 
In three aspects, the validation accuracy of MARE-hard was very close to BERT, and exceeded MARE-AMD by 3.5\%, 4.8\%, and 6.3\%, respectively. 
Meanwhile, MARE-hard leads MARE-AMD by 23.1\%, 23.4\%, and 78.1\% in rationalization performance, respectively. 
% The performance of soft deletion in the difficult aspect ("\textit {taste}" aspect) is very poor. 
The reason is that AMD fails to effectively separate tokens corresponding to different aspects, leading to information leakage and hindering accurate rationale extraction.
This indicates that AMD is not suitable for multi-aspect collaborative coding, and also proves the necessity and effectiveness of using hard deletion.

\begin{table}[!htpb]
    \centering
    \resizebox{0.48\textwidth}{!}{
    \begin{tabular}{llclclc}
    \toprule
    \multicolumn{1}{l}{\multirow{2}{*}{Methods}} & \multicolumn{2}{c}{Appearance}                             & \multicolumn{2}{c}{Aroma}                             & \multicolumn{2}{c}{Palate}                             \\\cline{2-7}
    \multicolumn{1}{c}{}                    & \multicolumn{1}{c}{ValAcc} & F1                    & \multicolumn{1}{c}{ValAcc} & F1                    & \multicolumn{1}{c}{ValAcc} & F1                    \\\hline 
    BERT                                    & \textbf{90.2}                       & \multicolumn{1}{l}{-} & \textbf{89.5}                       & \multicolumn{1}{l}{-} & \textbf{86.8}                       & \multicolumn{1}{l}{-} \\
    MARE-AMD                                     & 85.7                       & 69.4                  & 84.3                       & 68.6                  & 78.4                       & 3.9                   \\
    MARE-hard                                     & 89.2                       & \textbf{92.5}         & 89.1                       & \textbf{92.0}         & 84.7                       & \textbf{82.0}      \\\bottomrule  
    \end{tabular}
    }
    \caption{Ablation study on different delete methods.}
    \label{table:deletion}
    % \vspace{-1em}
\end{table}

\subsubsection{Special Token Initialization}

% To measure the impact of the initialization method of special tokens on the performance of MARE, this section measures three initialization methods:
To evaluate the impact of different initialization methods for special tokens on the model performance, this section explores three distinct initialization approaches:
\begin {itemize}
\item random initialization: The first special token is initialized by [CLS], while all other special tokens are randomly initialized.
\item CLS initialization: All the special tokens are initialized by [CLS].
\item sharing initialization: All the special tokens are shared and initialized by [CLS].
\end {itemize}

The performance comparisons are shown in Table~\ref{table:init}.
MARE-CLS is slightly better than MARE-random and the MARE-share performs the worst. 
We found that MARE share cannot distinguish the differences in sparsity between different aspects.
MARE-CLS achieves the best performance because the special token [CLS] is a highly informative embedding after pre-training.
By default, MARE uses the CLS initialization.

\begin{table}[!htpb]
    \centering
    \resizebox{0.48\textwidth}{!}{
    \begin{tabular}{llclclc}
    \toprule
    \multicolumn{1}{l}{\multirow{2}{*}{Methods}} & \multicolumn{2}{c}{Appearance}                             & \multicolumn{2}{c}{Aroma}                             & \multicolumn{2}{c}{Palate}                             \\\cline{2-7}
    \multicolumn{1}{c}{}                    & \multicolumn{1}{c}{ACC} & F1                    & \multicolumn{1}{c}{ACC} & F1                    & \multicolumn{1}{c}{ACC} & F1                    \\\hline 
    MARE-random                                     & \textbf{85.7}                       & 87.1                  & 85.4                       & 90.7                  & 87.0                       & 80.9                   \\
    MARE-share                                     & \textbf{85.7}                       & 85.1                  & 84.3                       & 88.1                  & 87.1                       & 79.0                   \\
    MARE-CLS                                     & 85.6                       & \textbf{92.5}         & \textbf{86.0}                       & \textbf{92.0}         & \textbf{88.0}                       & \textbf{82.0}      \\\bottomrule  
    \end{tabular}
    }
    \caption{Ablation study on different initialization strategies.}
    \label{table:init}
    % \vspace{-1em}
\end{table}

\section{Conclusion}
This paper proposed a Multi-Aspect Rationale Extractor to solve the limitations of traditional uni-aspect encoding models.
MARE can collaboratively predict and interpret multiple aspects of text simultaneously.
% Specifically, MARE introduces multiple independent special tokens and a hard deletion mechanism, allowing the model to independently capture information from multiple segments.
% At the same time, MARE introduces a multi-task training approach, which sequentially trains data from various aspects, saving a lot of training costs.
% Moreover, MARE shares the same BERT encoder across multiple aspects.
% Specifically, MARE introduced several independent special tokens to aggregate 
% the information of multiple segments separately. 
% Moreover, MARE employed a multi-aspect multi-head attention mechanism that allowed tokens to interact with each other within each segment while remaining independent across segments. 
Additionally, MARE incorporated multi-task training, sequentially training on data from each aspect, thereby significantly reducing training costs.
Extensive experimental results on two unsupervised rationale extraction datasets have shown that the rationalization performance of MARE is superior to all previous models.
Ablation studies further demonstrated the effectiveness of our method.

\section*{Limitations}
All of the above experiments have demonstrated the effectiveness of our method, but there are some limitations.
MARE needs to prepend some special tokens in front of the input, which increases the computational overhead.
Meanwhile, MARE can only adapted in encoder-based pre-trained language models. We are working hard to apply it to decoder-only models so that MARE can explain the predictions of LLMs. 
We will try to eliminate these limitations in our future work.

\section*{Acknowledgements}
This work was supported in part by the National Key Research and Development Program of China (No.2021YFF1201200), the Science and Technology Major Project of Changsha (No.kh2402004). This work was carried out in part using computing resources at the High-Performance Computing Center of Central South University.

% This document has been adapted by Yue Zhang, Ryan Cotterell, and Lea Frermann from the style files used for earlier ACL and NAACL proceedings, including those for 
% ACL 2020 by Steven Bethard, Ryan Cotterell, and Rui Yan,
% ACL 2019 by Douwe Kiela and Ivan Vuli\'{c},
% NAACL 2019 by Stephanie Lukin and Alla Roskovskaya, 
% ACL 2018 by Shay Cohen, Kevin Gimpel, and Wei Lu, 
% NAACL 2018 by Margaret Mitchell and Stephanie Lukin,
% Bib\TeX{} suggestions for (NA)ACL 2017/2018 from Jason Eisner,
% ACL 2017 by Dan Gildea and Min-Yen Kan, NAACL 2017 by Margaret Mitchell, 
% ACL 2012 by Maggie Li and Michael White, 
% ACL 2010 by Jing-Shin Chang and Philipp Koehn, 
% ACL 2008 by Johanna D. Moore, Simone Teufel, James Allan, and Sadaoki Furui, 
% ACL 2005 by Hwee Tou Ng and Kemal Oflazer, 
% ACL 2002 by Eugene Charniak and Dekang Lin, 
% and earlier ACL and EACL formats written by several people, including
% John Chen, Henry S. Thompson and Donald Walker.
% Additional elements were taken from the formatting instructions of the \emph{International Joint Conference on Artificial Intelligence} and the \emph{Conference on Computer Vision and Pattern Recognition}.

\bibliography{reference}
\bibliographystyle{acl_natbib}

\appendix

\section{Implementation Details}
\subsection{Main Experiments}
% \label{sec:imp_main}
% \paragraph{settings}
\label{sec:setting2}
In the experiment, we utilize the Pytorch~\cite{pytorch} deep learning framework and the huggingface transformers library~\cite{wolf2019huggingface} to implement MARE. BERT~\cite{JacobDevlin2018BERTPO} will be deployed as the backbone in MARE. MARE uses the AdamW optimizer~\cite{AdamW} to optimize parameters, with a learning rate set to $3 \times 10^{-5}$ and a weight decay set to $0.0$.
To control the sparsity and continuity of the generated rationales, this paper applies the "Cliff" deletion strategy, where the $k$ is fixed at $9$. In addition, we use grid search to select the most suitable hyperparameters $\beta$ and $\gamma$ from the candidate set $\{0.7, 1, 3, 5, 7 \}$.
We assume $\beta=\gamma$ in our experiments and select $\beta=\gamma=[0.7, 3, 3]$ for the BeerAdvocate dataset and Hotel Review dataset, respectively.
During the training process, we adopt a balanced round-robin method to iteratively sample data from all aspects. Set the batch size to 64 and limit the maximum sequence length to 256. For the BeerAdvocate dataset, MARE was trained for 15 epochs. However, considering the large scale of the Hotel Review dataset, the model only iteratively trained 5 epochs.

\subsection{Implementation Details of Token Deletion}
After obtaining the mask $\tilde{M}$ and its binarized counterpart $\tilde{\mathbf{M}}$ shown in Equation~\ref{eq:ex_dot} and \ref{eq:binarize}, we indeed multiply it with the attention score matrix to implement the token deletion. Specifically, the implementation in PyTorch is as follows:
\begin{lstlisting}[language=Python, caption={Token Deletion}, numbers=none]
def multi_head_attention(..., M, M_):
    ...

    att_score = Q @ K.transpose(-1, -2) / math.sqrt(d_head)

    # for token deletion
    M_grad = M + M_ - M_.detach()
    deleted_att_score = M_grad * torch.softmax(att_score, dim=-1)
    
    return deleted_att_score @ V
\end{lstlisting}
This implementation achieves the following:
\begin{itemize}
    \item The binary mask $\tilde{\mathbf{M}}$ determines which token pairs can interact (value 1) and which cannot (value 0).
    \item Multiplying $\mathbf{M\_grad}$ with the attention score matrix ($att\_score$) effectively zeroes out attention scores between tokens of different aspects.
    \item The resulting attention scores are then used to compute the weighted sum of value vectors ($\mathbf{V}$).
\end{itemize}
This approach ensures that tokens within the same aspect can interact through the attention mechanism, while interactions between tokens of different aspects are prevented. This aligns with our goal of allowing aspect-specific information to be aggregated separately.

\section{Cliff Decay}
\label{appendix:cliff_decay}
The Cliff decay strategy is defined as follows:
\begin{itemize}
    \item For layers $i < x$: All tokens are retained.
    \item For layers $i \ge x$: A proportion $p$ of tokens are deleted.
\end{itemize}
Here, $x$ is the layer at which deletion begins, and $p$ is the deletion proportion. In our experiments, we set $x = 9$, with $p$ varying by aspect.

\section{Training Stability of MARE}
\label{appendix:stability}
We have conducted additional experiments with different seeds to validate our training stability. 
Table~\ref{tab:std} shows the standard deviations of F1 scores across 3 different seeds.
As shown in the table, MARE demonstrates good stability, particularly in the Appearance and Aroma aspects. We believe this stability is partly due to the multi-aspect nature of our model, which allows it to leverage internal correlations between different aspects.
\begin{table}[!htpb]
\centering
\begin{tabular}{c|ccc}
\toprule
Method & Appearance & Aroma & Palate \\ \hline
MARE   & 0.4        & 0.3   & 1.3 \\ \bottomrule
\end{tabular}
\caption{The F1 standard deviations of MARE across 3 different seeds.}
\label{tab:std}
\end{table}

\section{Experiments on the Correlated BeerAdvocate Dataset}
To evaluate the impact of spurious correlation on model performance, we also conduct experiments on the correlated BeerAdvocate dataset~\cite{beer}.
The overall performance is shown in Table~\ref{table:yofov2_correlate}.
As we can see, MARE achieves state-of-the-art performance and is better than existing methods for a large margin.
We attribute this to the effectiveness of collaborative coding, demonstrating that internal correlations can suppress spurious correlations.

\begin{table*}[]
    % \footnotesize
    \centering
    \resizebox{\textwidth}{!}
    {
    \begin{tabular}{l|l|llll|llll|llll}
    \toprule
    {\multirow{2}{*}{Methods}} & \multicolumn{1}{c|}{\multirow{2}{*}{S}} & \multicolumn{4}{c|}{Appearance}                                                                   & \multicolumn{4}{c|}{Aroma}                                                                        & \multicolumn{4}{c}{Palate}                                                                       \\\cline{3-14} 
    \multicolumn{1}{c|}{}                         & \multicolumn{1}{c|}{}                   & \multicolumn{1}{c}{ACC} & \multicolumn{1}{c}{P} & \multicolumn{1}{c}{R} & \multicolumn{1}{c|}{F1} & \multicolumn{1}{c}{ACC} & \multicolumn{1}{c}{P} & \multicolumn{1}{c}{R} & \multicolumn{1}{c|}{F1} & \multicolumn{1}{c}{ACC} & \multicolumn{1}{c}{P} & \multicolumn{1}{c}{R} & \multicolumn{1}{c}{F1} \\\hline
    RNP\cite{lei-etal-2016-rationalizing}                                           & \multirow{6}{*}{10}                     & -                       & 32.4                  & 18.6                  & 23.6                    & -                       & 44.8                  & 32.4                  & 37.6                    & -                       & 24.6                  & 23.5                  & 24.0                     \\
    HardKuma\cite{bastings-etal-2019-interpretable}                                      &                                         & -                       & 53.6                  & 28.7                  & 37.4                    & -                       & 29.3                  & 25.9                  & 27.3                    & -                       & 7.7                   & 6.0                     & 6.8                    \\
    INVRAT\cite{INVRAT}                                        &                                         & -                       & 42.6                  & 31.5                  & 36.2                    & -                       & 41.2                  & 39.1                  & 40.1                    & -                       & 34.9                  & 45.6                  & 39.5                   \\
    Inter-RAT\cite{inter-rat}                                     &                                         & -                       & 66.0                    & 46.5                  & 54.6                    & -                       & 55.4                  & 47.5                  & 51.1                    & -                       & 34.6                  & 48.2                  & 40.2                   \\
    MGR\cite{MGR}                                           &                                         & 80.5                    & 87.5                  & 51.7                  & 65.0                      & 89.7                       & 78.7                  & 52.2                  & 62.8                    & 86.0                      & 65.6         & 57.1                  & 61.1                   \\
    YOFO~\cite{YOFO}                                    &                                         & {87.7}           & {96.4}         & {61.9}         & {75.4}           & \textbf{92.7}           & {95.4}         & {65.2}         & {77.5}           & \textbf{91.9}           & {67.4}                  & {67.4}         & {67.4}          \\
    MARE(\textit{ours})                                    &                                         & \textbf{88.9}           & \textbf{99.0}         & \textbf{62.2}         & \textbf{76.4}           & {92.0}           & \textbf{97.5}         & \textbf{66.2}         & \textbf{78.9}           & {90.8}           & \textbf{81.6}                  & \textbf{72.8}         & \textbf{77.0}          \\\hline
    RNP\cite{lei-etal-2016-rationalizing}                                           & \multirow{6}{*}{20}                     & -                       & 39.4                  & 44.9                  & 42.0                      & -                       & 37.5                  & 51.9                  & 43.5                    & -                       & 21.6                  & 38.9                  & 27.8                   \\
    HardKuma\cite{bastings-etal-2019-interpretable}                                      &                                         & -                       & 64.9                  & 69.2                  & 67.0                      & -                       & 37.0                    & 55.8                  & 44.5                    & -                       & 14.6                  & 22.3                  & 17.7                   \\
    INVRAT\cite{INVRAT}                                        &                                         & -                       & 58.9                  & 67.2                  & 62.8                    & -                       & 29.3                  & 52.1                  & 37.5                    & -                       & 24.0                    & 55.2                  & 33.5                   \\
    Inter-RAT\cite{inter-rat}                                     &                                         & -                       & 62.0                    & 76.7                  & 68.6                    & -                       & 44.2                  & 65.4                  & 52.8                    & -                       & 26.3                  & 59.1                  & 36.4                   \\
    MGR\cite{MGR}                                           &                                         & 85.6                    & 76.3         & 83.6                  & 79.8                    & 89.6                    & 64.4                  & 81.3                  & 71.9                    & 89.3                    & {47.1}         & 73.1                  & {57.3}          \\
    YOFO~\cite{YOFO}                                    &                                         & {88.4}           & {77.5}                  & {87.6}         & {82.2}             & {91.9}           & \textbf{78.7}         & {92.8}         & \textbf{85.2}           & {91.3}           & 44.6                    & {75.4}         & 56.0                   \\
    MARE(\textit{ours})                                    &                                         & \textbf{90.6}           & \textbf{81.4}                  & \textbf{92.4}         & \textbf{86.6}             & \textbf{92.1}           & {74.0}         & \textbf{95.0}         & {83.2}           & \textbf{91.9}           & \textbf{47.3}                    & \textbf{88.0}         & \textbf{61.5}                   \\\hline
    RNP\cite{lei-etal-2016-rationalizing}                                           & \multirow{6}{*}{30}                     & -                       & 24.2                  & 41.2                  & 30.5                    & -                       & 27.1                  & 55.7                  & 36.4                    & -                       & 15.4                  & 42.2                  & 22.6                   \\
    HardKuma\cite{bastings-etal-2019-interpretable}                                      &                                         & -                       & 42.1                  & 82.4                  & 55.7                    & -                       & 24.6                  & 57.7                  & 34.5                    & -                       & 21.7                  & 49.7                  & 30.2                   \\
    INVRAT\cite{INVRAT}                                        &                                         & -                       & 41.5                  & 74.8                  & 53.4                    & -                       & 22.8                  & 65.1                  & 33.8                    & -                       & 20.9                  & 71.6                  & 32.3                   \\
    Inter-RAT\cite{inter-rat}                                     &                                         & -                       & 48.1                  & 82.7                  & 60.8                    & -                       & 37.9                  & 72.0                    & 49.6                    & -                       & 21.8                  & 66.1                  & 32.8                   \\
    MGR\cite{MGR}                                           &                                         & 88.5                    & 57.2                  & 93.9         & 71.1                    & 91.6                    & 45.8                  & 87.4         & 60.1                    & 89.3                    & 27.3                  & 66.5                  & 38.7                   \\
    YOFO~\cite{YOFO}                                    &                                         & \textbf{88.9}           & {63.5}         & {94.3}                  & {75.9}           & {92.4}           & {53.6}         & {88.7}                  & {66.8}           & {91.6}           & {34.0}         & {75.7}         & {46.9}         \\
    MARE(\textit{ours})                                    &                                         & {88.7}           & \textbf{65.9}         & \textbf{96.8}                  & \textbf{78.4}           & \textbf{92.9}           & \textbf{55.2}         & \textbf{91.9}                  & \textbf{69.0}           & \textbf{92.7}           & \textbf{35.7}         & \textbf{79.0}         & \textbf{49.2}         \\
    \bottomrule
    \end{tabular}
    }
    \caption{The results of different methods on correlated BeerAdvocate Dataset~\cite{beer}.
    %“*”和“**”表示从获得的结果\cite{dr}and~\cite{MGR}。\textbf{粗体}表示不同方法中的最佳结果。\underline{下划线}表示不同方法中的第二名结果。The results of different Methods on correlated BeerAdvocate Dataset~\cite{beer}. "*" and "**" represent the results obtained from ~\cite{inter-rat} and ~\cite{MGR}, respectively. \textbf{Bold} indicates the best results among all Methods in the same setting. 
    }
    \label{table:yofov2_correlate}

\end{table*}

\end{document}